\pdfoutput=1

\documentclass[11pt]{article}

\usepackage{acl}

\usepackage{times}
\usepackage{latexsym}

\usepackage[T1]{fontenc}

\usepackage[utf8]{inputenc}

\usepackage{microtype}

\usepackage{enumitem}
\setitemize{noitemsep,topsep=0pt,parsep=0pt,partopsep=0pt}
\definecolor{cite}{rgb}{0.6,0.6,1.0}

\definecolor{todo}{rgb}{1,0.5,0}

\usepackage{booktabs}
\usepackage{graphicx}
\usepackage{float}
\usepackage{tabularx}
\newcommand{\corpus}{\textsc{NLPeer}}

\newcolumntype{L}[1]{>{\raggedright\arraybackslash}m{#1}}
\newcolumntype{C}[1]{>{\centering\arraybackslash}m{#1}}
\newcolumntype{R}[1]{>{\raggedleft\arraybackslash}m{#1}}

\title{\corpus: A Unified Resource for the Computational Study of Peer Review}

\author{Nils Dycke, Ilia Kuznetsov, Iryna Gurevych\\ 
Ubiquitous Knowledge Processing Lab (UKP Lab)\\
Department of Computer Science and Hessian Center for AI (hessian.AI)\\
Technical University of Darmstadt \\
\texttt{ukp.informatik.tu-darmstadt.de}}

\begin{document}

\maketitle
\begin{abstract}
Peer review constitutes a core component of scholarly publishing; yet it demands substantial expertise and training, and is susceptible to errors and biases. Various applications of NLP for peer reviewing assistance aim to support reviewers in this complex process, but the lack of clearly licensed datasets and multi-domain corpora prevent the systematic study of NLP for peer review. To remedy this, we introduce \corpus\ -- the first ethically sourced multidomain corpus of more than 5k papers and 11k review reports from five different venues. In addition to the new datasets of paper drafts, camera-ready versions and peer reviews from the NLP community, we establish a unified data representation and augment previous peer review datasets to include parsed and structured paper representations, rich metadata and versioning information. We complement our resource with implementations and analysis of three reviewing assistance tasks, including a novel guided skimming task.
Our work paves the path towards systematic, multi-faceted, evidence-based study of peer review in NLP and beyond. The data\footnote{\url{https://tudatalib.ulb.tu-darmstadt.de/handle/tudatalib/3618}} and code\footnote{\url{https://github.com/UKPLab/nlpeer}} are publicly available. 
\end{abstract}

\section{Introduction}

Research publication is the primary unit of scientific communication. To ensure publication quality and to prioritise research outputs, most scientific communities rely on peer review \cite{johnson_stm_2018} -- a distributed procedure where independent referees determine if a manuscript adheres to the standards of the field. Despite its utility and wide application, peer review is an effortful activity that requires expertise and is prone to bias \cite{bias1, lee_bias_2013, resub}. An active line of research in NLP for peer review strive to address these challenges by supporting the underlying editorial process \cite[e.g.][]{price_computational_2017, kang2018dataset, shah_principled_2019}, decision making \cite[e.g.][]{shen-etal-2022-mred, dycke2021ranking, ghosal_deepsentipeer_2019}, review writing \cite{reviewadvisor}, and by studying review discourse \cite[e.g.][]{kennard-etal-2022-disapere, kuznetsov2022revise, ampere, ape, hedgepeer}.

\begin{figure}[t!] 
	\centering
	\includegraphics[width=0.95\linewidth]{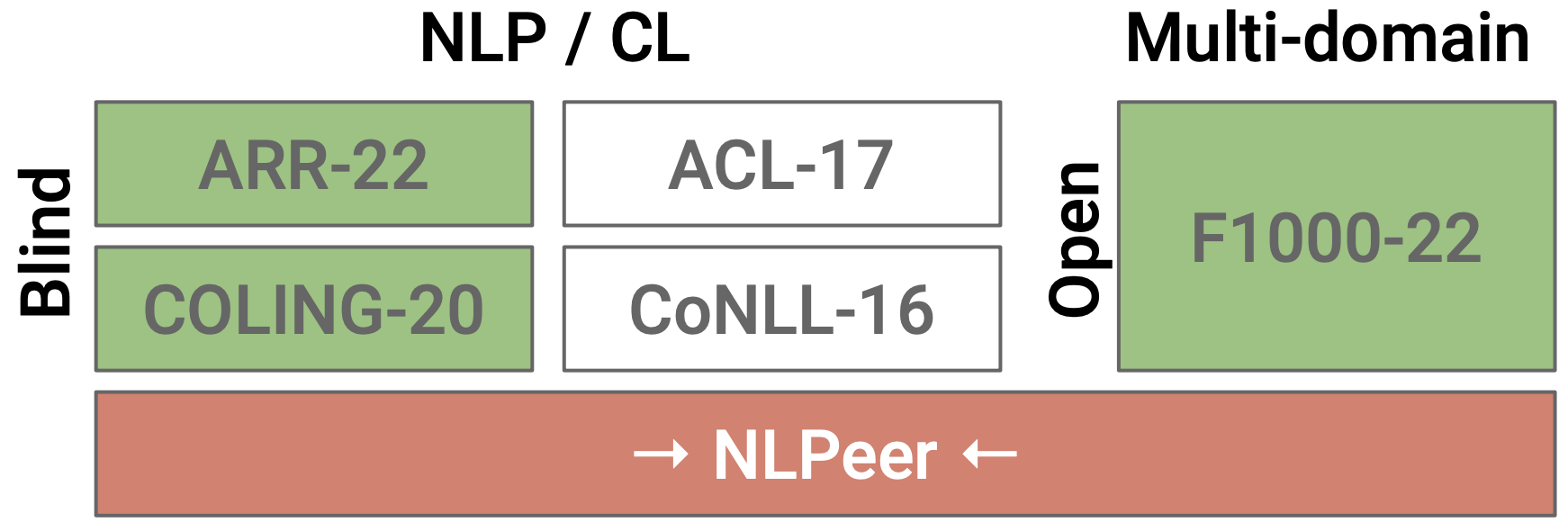}
	\caption{\corpus\ unites openly licensed datasets from different research communities, reviewing systems and time periods, including three previously unreleased text collections: ARR-22, COLING-20 and F1000-22.}
\label{fig:nlpeer_structure}
\end{figure}

Despite the methodological advancements, several factors prevent NLP research for peer review at large. The computational study of peer review lacks a (1) \textbf{solid data foundation}: reviewing data is rarely public and comes with legal and ethical challenges; existing sources of peer reviewing data and the derivative datasets are not licensed, which legally prevents reuse and redistribution \cite{workflow_3y}. Peer reviewing practices vary across research communities \cite{walker_emerging_2015, bornmann_scientific_2011} -- yet the vast majority of NLP research in peer review so far focused on a few machine learning conferences that make their data available through the OpenReview.net platform \cite[e.g.][]{kennard-etal-2022-disapere,shen-etal-2022-mred}. A (2) \textbf{multi-domain perspective on peer review} is thus missing, and the transferability of findings between different communities and reviewing workflows remains unclear. Finally, a (3) \textbf{unified data model} for representing peer reviewing data is lacking: most existing datasets of peer reviews adhere to task-specific data models and formats, making it hard to develop and evaluate approaches for peer review support across datasets and domains.

To address these issues, we introduce \corpus. 
We apply a state-of-the-art workflow \cite{workflow_3y} to gather ethically and legally compliant reviewing data from natural language processing (NLP) and computational linguistics (CL) communities. We complement it with multi-domain reviewing data from the F1000 Research\footnote{\url{https://f1000research.com/}} platform and historical data from the openly licensed portion of the PeerRead \cite{kang2018dataset} corpus. The resulting resource (Figure \ref{fig:nlpeer_structure}) is the most comprehensive collection of clearly licensed, open peer reviewing datasets available to NLP to date.

\corpus\ includes peer reviews, paper drafts and revisions from diverse research fields and reviewing systems, over the time span from as early as 2012 until 2022. This -- for the first time -- enables systematic computational study of peer review across domains, communities, reviewing systems and time. The paper revisions make \corpus\ well-suited for the study of collaborative text work. 

To facilitate the analysis, we unify the datasets under a common data model that preserves document structure, non-textual elements and is well suited for cross-document analysis. To explore the new possibilities opened by our resource, we conduct cross-domain experiments on \textit{review score prediction} (to encourage consistent review scores), \textit{pragmatic labeling} (to encourage balanced reviews) and the novel \emph{guided skimming for peer review} task (to help guide review focus), along with easy-to-extend implementations. Our results indicate substantial variation in performance of NLP assistance between venues and research communities, point at synergies between different approaches to review structure analysis, and pave the path towards exploiting cross-document links between peer reviews and research papers for language model benchmarking.

In summary, this work contributes (1) the first unified, openly licensed, multi-domain collection of datasets for the computational study of peer review, including (2) two novel datasets of peer reviews from the NLP and CL communities, and complemented by a (3) descriptive analysis of the resulting data and (4) extensive experiments in three applied NLP tasks for peer reviewing assistance.

\section{Background}

\subsection{Peer Reviewing Terminology} 

During peer review, authors submit their \emph{paper}\footnote{We use the terms manuscript, paper  etc. interchangeably.} to the editors, often via a peer reviewing platform. The manuscript is distributed among reviewers who produce reviewing reports -- or \emph{reviews} -- evaluating the submission. As a result, the submission might be accepted, rejected or further adjusted, producing a \emph{revision}. Some reviewing systems allow additional exchange, e.g.\ author responses and meta-reviews. However, here we focus on papers, reviews and revisions, which constitute the core of \emph{peer reviewing data}. 

There exist different implementations of peer review, including \emph{blind review} (where reviewer and/or author identities are hidden to promote objectivity) and \emph{open review} (identities are open). Moreover, reviewing standards and practices vary by research community, venue and publication type. Review forms and templates are one important varying factor; in the following, we differentiate between \textit{unstructured} (only one main text) and \textit{structured} (several predefined sections) review forms. We refer to the particular implementation of peer review at a certain venue as \emph{reviewing system}. Along with the natural domain shift based on research field, the reviewing system contributes to the composition of the peer reviewing data it produces.

\subsection{Existing Data}

Peer review is a hard, time-consuming, subjective task prone to bias, and an active line of work in NLP aims to mitigate these issues. NLP for peer review crucially depends on the availability of peer reviewing data -- yet open data is scarce. The majority of existing NLP studies on peer review \cite[e.g.][]{kang2018dataset, reviewadvisor, ampere, kennard-etal-2022-disapere, ape, ghosal2022peer, shen-etal-2022-mred} draw their data from a few machine learning conferences on the OpenReview.net platform which -- at the time of writing -- does not attach explicit licenses to the publicly available materials, making reuse of the derivative data problematic. In addition, over-focusing on few selected conferences in machine learning limits the utility of NLP approaches to peer review for other areas of science and reviewing systems, and leaves the question of cross-domain applicability open. While the recent F1000RD corpus \cite{kuznetsov2022revise} addresses some of these issues, it lacks data from NLP and CS communities, and does not contain blind reviewing data, although single- and double-blind review are arguably standard in most research fields \cite{johnson_stm_2018}.

Ethical, copyright- and confidentiality-aware collection of peer reviewing data \cite{workflow_3y}, and transformation of this data into unified, research-ready datasets, require major effort. The purpose of \corpus\ is to provide the NLP community a head-start in ethically sound, cross-domain and cross-temporal study of peer review.

\subsection{NLP Approaches to Aiding Peer Review}

Recent years have seen a surge in NLP approaches to aid peer review; ranging from early works studying reviewing scores \cite{kang2018dataset} to more recent approaches attempting to align author responses to review reports automatically \cite{kennard-etal-2022-disapere}. 
The goal of our work is to provide diverse and clearly licensed source data to support the future studies in NLP for peer review assistance.

In addition, we explore the potential for cross-domain NLP-based peer reviewing assistance on three tasks detailed in Section \ref{sec:assistance}. Review score prediction has been first introduced in \cite{kang2018dataset} and further explored in \cite{ghosal_deepsentipeer_2019}. While following a similar setting, our study contributes to this line of research by exploring the task of score prediction across domains and research communities and provides new insights on the factors that impact the transferability of the task between reviewing systems. Pragmatic labeling has been previously explored in \cite{ampere, kuznetsov2022revise, kennard-etal-2022-disapere} and is usually cast as a discourse labeling task on free-form peer review text. Yet, many reviewing systems enforce the same discourse structure by employing structured peer review forms. Complementing the prior efforts, we explore the potential synergies between the two approaches. The guided skimming for peer review task is novel and builds upon the recent work in cross-document modeling for peer review by \citet{kuznetsov2022revise} and other related works \cite{qin2022structure,ghosal2022peer}.

\pagebreak
\section{\corpus}

\begin{figure}[t!] 
	\centering
	\includegraphics[width=0.95\linewidth]{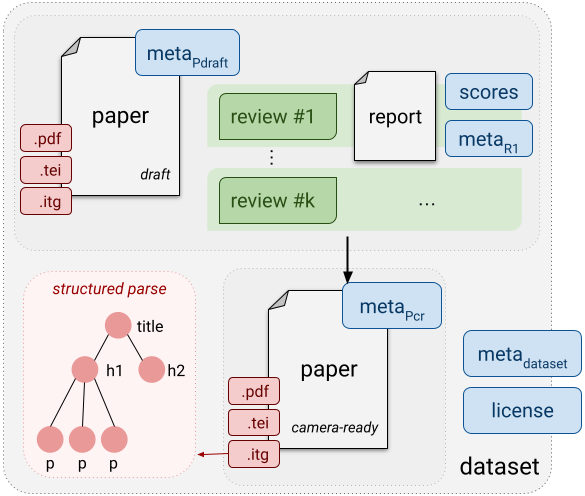}
	\caption{Unified data structure of \corpus.}
\label{fig:corpus_structure}
\end{figure}

\begin{table*}[t]
	\centering
    \resizebox{\linewidth}{!}{
	\begin{tabular}{R{2.8cm}C{1.7cm}C{2.1cm}C{1.7cm}C{1.9cm}C{1.8cm}C{1.1cm}}
	    \toprule
	    & \multicolumn{2}{C{3.6cm}}{{New Datasets}} & \multicolumn{3}{C{6cm}}{{Existing Datasets}} & \\
	    
	    \cmidrule(lr){2-3} \cmidrule(lr){4-6} %
	    
	    & \textbf{ARR-22} & \textbf{COLING-20} & \textbf{ACL-17} & \textbf{CONLL-16} & \textbf{F1000-22} & \textbf{total}\\
	    \midrule
	    reviewing system & blind & blind & blind & blind & open & both \\
	    domain & NLP/CL & NLP/CL & NLP/CL & NLP/CL & multi & multi \\
	    time (20XX) & 21-22 & 20 & 16-17 & 15-16 & 12-22 & 12-22 \\
	    \midrule
	    \# papers & $476$ & $89$ & $136$ & $22$ & $4949$ & $5672$\\
	    \% accepted & $100\%$ & $93\%$ & $67\%$ & $50\%$ & $34\%$ & $69\%^*$\\
        \# reviews & $684$ & $112$ & $272$ & $39$ & $10418$ & $11515$\\
        reviews per paper & $1.43\pm1.1$& $1.27\pm0.5$ & $2.0 \pm 0.8 $ & $1.77 \pm 0.6$ & $2.11 \pm 0.71$ & $1.72^*$\\
        \midrule
        \#sent. per paper & $220 \pm 62.9 $ & $195 \pm 47.9$ & $208 \pm 42.3$ & $199 \pm 35.6 $ & $158 \pm 90.3$ & $196^*$\\
        \#sent. per review & $31 \pm 27.0 $ & $30 \pm 22.8$ & $41 \pm 24.6 $ & $32 \pm 12.4 $ & $34 \pm 27.6$ & $32.2^*$ \\
		total \#tok reviews & $266$k & $45$k & $100$k & $16$k & $3.8$M & $4.2$M \\
		\bottomrule
	\end{tabular} 
    }
	\caption{For the datasets in \corpus, we report the mean and standard deviation of per-paper and per-review statistics. We report \% of accepted papers; for F1000-22 this is the \% of version one drafts with unanimous scores of "accept", as acceptance conceptually does not exist here. %
	Total statistics are summed and averaged ($^*$), respectively.}
	\label{t:datastats}
\end{table*}

\subsection{Datasets}

\corpus\ consists of five datasets: two datasets from previous work, two entirely new datasets from the NLP domain, as well as an up-to-date crawl of the F1000Research platform.

\paragraph{Prior datasets} Parts of the PeerRead data \cite{kang2018dataset} have been created with explicit consent for publication and processing by both reviewers and authors, and we include them into our resource.  \textbf{ACL-17}\footnote{55th Annual Meeting of the Association for Computational Linguistics (ACL)} and  \textbf{CONLL-16}\footnote{Conference on Computational Language Learning 2016} contain peer reviews and papers from the NLP domain, stem from a double-blind reviewing process, and use unstructured review forms with a range of numerical scores, e.g. substance and soundness. The data is licensed under CC-BY.%

\paragraph{F1000-22} is collected from F1000Research -- a publishing platform with an open post-publication reviewing workflow. Unlike other datasets in \corpus, F1000-22 covers a wide range of research communities from scientific policy research to medicine and public health. %
The reviewing process at F1000Research is fully open, with reviewer and author identities known throughout the process, contributing to the diversity of \corpus. F1000Research uses unstructured peer reviewing forms coupled with a single 3-point overall score (approve, reject, approve-with-reservations). The paper and review data are distributed under CC-BY license, which we preserve.

\paragraph{COLING-20 (New)} was collected via a donation-based workflow at the 28th International Conference on Computational Linguistics, in the NLP and computational linguistics domain. The data stems from a double-blind reviewing process; review forms include free-form report texts and multiple numerical scores, e.g. relevance and substance. We release this data under the CC-BY-NC-SA 4.0 license.%

\paragraph{ARR-22 (New)} was collected via the donation-based workflow proposed by \newcite{workflow_3y} at ACL Rolling Review (ARR) -- a centralized system of the Association for Computational Linguistics. \corpus\ includes peer reviewing data for papers later accepted at two major NLP venues -- ACL 2022\footnote{60th Annual Meeting of the ACL} and the NAACL 2022\footnote{2022 Annual Meeting of the NAACL} -- covering submissions to ARR from September 2021 to January 2022. The reviewing process at ARR is double-blind and uses standartized structured review forms that include strengths and weakness sections, overall and reproducibility scores, etc. We release this data under the CC-BY-NC-SA 4.0 license.

\subsection{Unification}

The diverse source datasets in \corpus\ were cast into a unified data model (Figure \ref{fig:corpus_structure}). Each paper is represented by the submission version and a revised camera-ready version, and is associated with one or more review reports. In the case of F1000-22 all revisions and their reviews are present. To unify the papers, we converted all drafts and revisions in \corpus\ into intertextual graphs (ITG) -- a recently proposed general document representation that preserves document structure, cross-document links and layout information \cite{kuznetsov2022revise}. We extended the existing ITG parser\footnote{\url{https://github.com/UKPLab/intertext-graph}} and combined it with GROBID \cite{GROBID} to process PDF documents and preserve line number information whenever available. Papers were supplemented with the PDF, XML and TEI source whenever available. Reviews were converted into a standardized format that accommodates structured and free-text reviews and arbitrary sets of scores. As paper revisions were not collected for some of the \corpus\ datasets, we have complemented existing data with camera-ready versions obtained via the ACL Anthology\footnote{\url{https://aclanthology.org/}}. Papers, reviews and datasets are accompanied with metadata, e.g. paper track information and licenses for individual dataset items. Further dataset creation details are provided in the Appendix \ref{as:corpus}.

\subsection{Ethics, Licensing and Personal Data} All datasets included in \corpus\ are distributed under an open Creative Commons license and were collected based on explicit consent, or an open license attached to the source data. The ARR-22 data collection process allowed reviewers to explicitly request attribution, and this information is included in the dataset. F1000Research uses open reviewer and author identities throughout the reviewing process; this information is preserved in F1000-22. Finally, the authors of camera-ready publications included in \corpus\ are attributed. In all other cases, review reports and paper drafts throughout \corpus\ are stripped of personal metadata; in addition, the reviews of all datasets apart from F1000-22 have been manually verified by at least one expert from our team to not contain personal information. Due to the practices of scientific publishing in the selected communities, \corpus\ only includes texts written in English.

\section{Statistics}

The datasets in \corpus\ originate from different domains and peer reviewing systems, each with its own policies, reviewing guidelines, and community norms. While a comprehensive comparison of publishing and reviewing practices lies beyond the scope of this work, this section provides a brief overview of the key statistics for \corpus\ datasets.

\paragraph{Documents and Texts}
Table \ref{t:datastats} reports the textual statistics of \corpus, comprising more than 4M peer review tokens in total, with more than 400K review tokens in the NLP/CL domain. The resource is diverse, and we observe high variability in review and paper composition among the datasets. For example, papers in F1000-22 come from a wide range of domains and cover a wide range of article types from case reports to position papers, reflected in the low number of sentences per paper ($158$) with high variance($\pm90.3$) compared to the other, NLP-based datasets with roughly $200$ sentences per paper and lower variance. Yet, we note that review lengths exhibit smaller variance across the datasets. Every paper in \corpus\ is associated with at least one review and one revision, making it suitable for the study of cross-document relationships.

\begin{figure} 
	\centering
	\includegraphics[width=\linewidth]{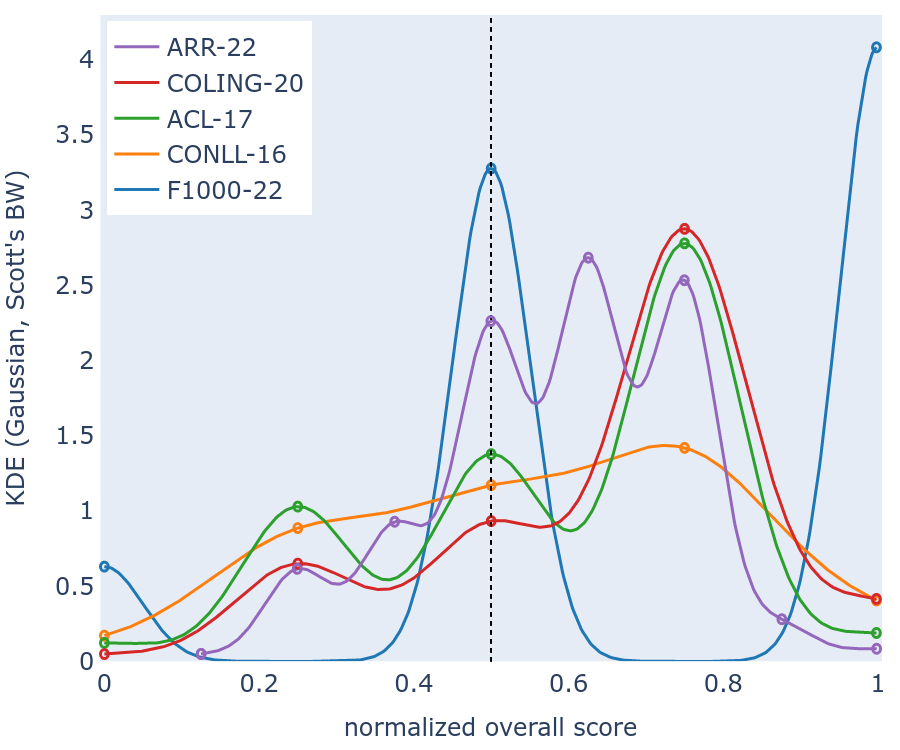}
	\caption{Normalized overall score per dataset (KDE estimate) with "borderline" (dotted).}
\label{fig:overall_scores}
\end{figure}

\paragraph{Scores and Acceptance}
The data collection workflow impacts the proportion of accepted papers in the dataset: from 100\% in ARR-22 which uses a strict confidentiality-aware collection procedure, to 34\% in F1000-22, %
where manuscripts are made available prior to peer reviewing and acceptance. Yet, acceptance per se is not an accurate proxy of review stance: a paper that is eventually accepted can receive critical reviews. To investigate, we turn to reviewing scores. Each of the reviewing systems in \corpus\ requires reviewers to assign a numerical rating to the papers. However, scoring scales and semantics differ across datasets: the NLP and CL conferences employ fine-grained scales (5- and 9-point), while F1000 Research uses a very coarse scale (3-point). Figure \ref{fig:overall_scores} shows the normalized distribution of overall scores for each dataset in \corpus. We observe that scores near the arguably most interesting region around the borderline to acceptance are well-represented, with a skew towards the positive end of the review score scale otherwise. For the donation-based datasets (all except F1000-22), this is likely due to participation bias \cite{workflow_3y}; for F1000-22, it is a result of the post-publication, revision-oriented reviewing workflow.  This fundamental difference between the peer reviewing systems in \corpus\ makes it an interesting and challenging target for the computational study of reviewing scores across reviewing systems and research communities.

\begin{figure} 
	\centering
	\includegraphics[width=0.95\linewidth]{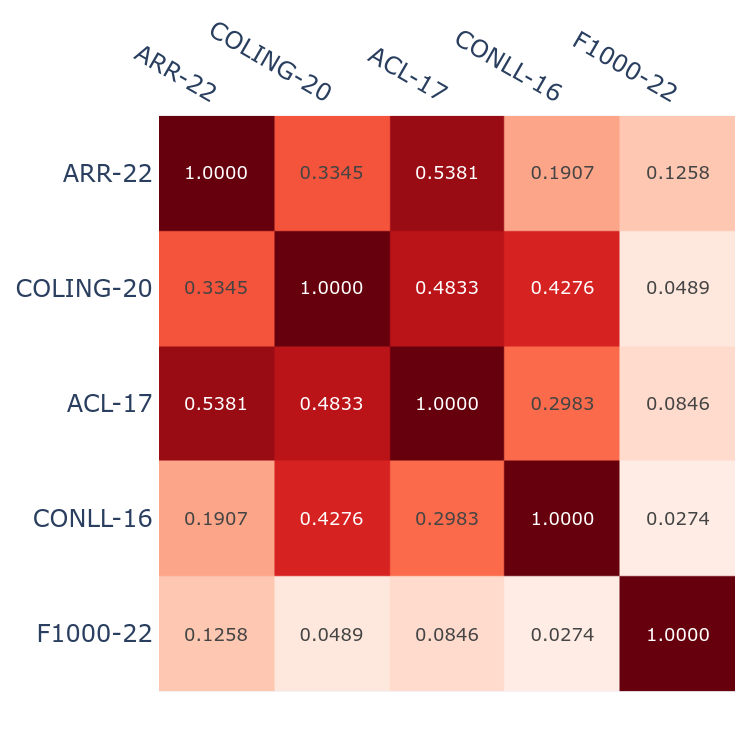}
	\caption{Vocabulary overlap based on the Jaccard metric on review lemmas of the datasets in \corpus.}
\label{fig:vocab_overlap}
\end{figure}

\paragraph{Domain Structure}
The vocabulary of a text collection is an important proxy for describing its language variety \cite{plank}, and higher vocabulary overlap indicates shared terminology and topical alignment between datasets. To investigate the domain structure of \corpus, we measure the vocabulary overlap of review texts based on the Jaccard metric for the top $10\%$ most frequent lemmas excluding stopwords similar to \newcite{zhang2021unsupervised}. As Figure \ref{fig:vocab_overlap} demonstrates, reviews from the NLP/CL communities (ARR-22, ACL-17, COLING-20) are most similar ($0.37$-$0.53$), while F1000-22 is most similar to ARR-22 with a notably lower score than the within-community comparison ($\Delta \approx 0.25$). This illustrates that despite domain differences, the review reports do share general wording (e.g. "model", "method", "author") that is independent from lower-frequency field-specific terminology (e.g. "cross-entropy", "feed-forward", "morpheme"). \corpus\ includes datasets with linguistically diverse review reports while maintaining a domain-independent base vocabulary characteristic for the genre of peer reviews.\footnote{The paper vocabulary overlap follows a similar trend, see Appendix (Figure \ref{afig:paper_vocab_overlap}).} This makes the investigation of cross-domain NLP for reviewing assistance a promising research avenue.

\section{Reviewing Assistance with \corpus}
\label{sec:assistance}

\begin{figure}
	\centering
	\includegraphics[width=0.95\linewidth]{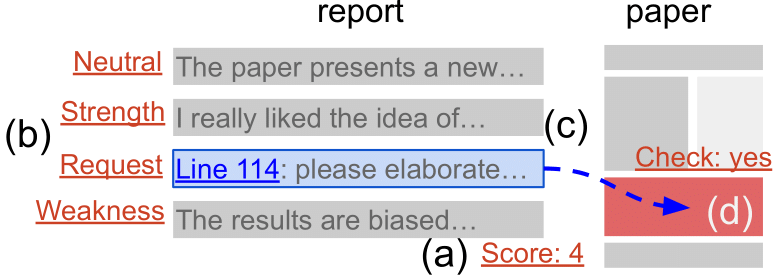}
	\caption{\corpus\ tasks: given a review and a paper, we (a) predict the review score, (b) predict pragmatic class per review sentence, and (c) use explicit links to (d) find review-focus paragraphs in new papers.}
\label{fig:tasks}
\end{figure}

\corpus\ is a unique resource for the study of computational assistance during peer review authoring. To demonstrate its versatility, we define three tasks -- review score prediction, pragmatic labeling and guided skimming -- that are anchored in realistic assistance scenarios and targeted towards helping junior reviewers to improve their review. Figure \ref{fig:tasks} illustrates the tasks. For readability, we use $s$ to denote sentences, $\textit{par}$ for paragraphs, and $\textit{sec}$ to denote sections for reviews (e.g. $s^R$) and papers (e.g. $s^P$), respectively. We report the main results here, and refer to the Appendix \ref{as:experiments} for details and the \nameref{s:limits} Section for an ethical discussion of risks and opportunities of these tasks.

\subsection{General Setup}

\paragraph{Baseline Models}
Our experiments aim to assess the difficulty of each assistance task across the subsets of \corpus. Hence, we base our experiments on well-established large language models (LLM) from the general and scientific domain -- RoBERTa \cite{liu2019roberta}, BioBERT \cite{lee2020biobert} and SciBERT \cite{beltagy2019scibert} -- with the recommended default hyperparameter settings.

\paragraph{Fine-tuning and Evaluation}
For each task, we split the data into a training (70\%), development (10\%) and test set (20\%), and fine-tune the pre-trained LLM on the training split using a task-specific prediction head. To account for small dataset sizes, %
we fine-tune each model with small learning rates ([$1$, $3\times10^{-5}$]) and a linear warm-up schedule, and allow training for up to $20$ epochs, following existing recommendations \cite{mosbach2020stability}. We repeat the experiments ten-times with different random seeds and report the median and standard deviation of each performance metric across the runs. %
All experiments were run on a single NVIDIA A100 GPU, summing to a total of roughly seven days computing time all together.

\subsection{Review Score Prediction}
Reviewer rating behavior is heterogeneous; the correspondence between review ratings and review texts often lacks consistency \cite{wang2018your}. The differences in rating assignment might exist both on the individual level (i.e. due to lack of experience) and on the community level (i.e. when reviewing for a large multi-track conference). Here, a review score prediction model can suggest a score to the reviewer that is typical for the given report and sub-community. This suggestion can serve as feedback to revise inconsistent scores.

\paragraph{Task.} Following prior work \cite{kang2018dataset, ghosal_deepsentipeer_2019,stappen2020uncertainty}, we cast review score prediction (RSP) as a regression task where given the review text in $R$ and the paper abstract $a^P$ the model should predict the overall review score $q^R$ mapped to the respective scale. We provide the paper abstract as an input to the model to allow to contextualize the review text; hereby, the model can resolve coreferences and weigh review statements in relation to the paper. We measure the performance by the mean root squared error (MRSE). As one key challenge of review scoring is the mapping of an overall assessment to a discrete score scale, we also measure the classification performance of the regression model in terms of the F1-score by splitting the real-valued outputs into equally sized intervals according to the respective review score scale. We highlight that this task framing does not account for specific score semantics, but in exchange permits direct model transfer and comparison across domains. We leave the in-detail study of RSP as a classification task, for instance using in-context learning \cite{brown2020language} on score semantic labels, to future work.

\paragraph{Setup.} For each dataset in \corpus, we consider all reviews of the initial paper draft; this means, for F1000-22 we include only the reviews of the first paper version. As the input, we concatenate the review report sections $\textit{sec}_1^R, \textit{sec}_2^R... \in R$ and the paper abstract $a^P$, which we separate by a special token. To account for the limitations of the LLMs, we truncate the resulting text to $512$ tokens. For training, we normalize the scores $q^R$ to the interval $[0, 1]$ considering the maximum and minimum scores of the respective rating scale. 
The same paper can receive multiple reviews; we ensure that all reviews of the same paper belong to the same data split. For representative sampling, we ensure that the distribution of reviews per paper is similar across splits (see Appendix \ref{as:experiments}). %

\paragraph{Results.}
Table \ref{t:rsp} summarizes score prediction results for the best-performing LLM by dataset. %
Neural models substantially outperform the mean score baseline for ARR-22 and F1000-22, yet for ACL-17, CONLL-16 and COLING-20 they perform on-par or worse. We note that the latter datasets also have the lowest number of samples. To investigate a potential connection, we have trained a RoBERTa model on $30\%$ of the ARR-22 training set, resulting in a median $0.44$ MRSE and $0.24$ F1-macro score on the test set -- on par with the mean baseline, and comparable to the similarly-sized ACL-17. This suggests that the dataset size has the strongest impact on score prediction performance, encouraging future  reviewing data collection efforts, as well as the study of few-shot transfer between high- and low-resource peer reviewing systems and domains. %

\begin{table}[t]
	\centering
 \resizebox{\linewidth}{!}{
	\begin{tabular}{R{2.3cm} C{2.2cm} C{2.2cm}}
	    \toprule
	     & \textbf{MRSE $\downarrow$} & \textbf{F1-macro $\uparrow$}\\
		\midrule
		\textbf{ARR-22$\dagger$} & $.45$ / \textbf{.37} $\pm .02$ & .24 / \textbf{.46} $\pm .04$ \\
		
		\textbf{COLING-20$\dagger$} & \textbf{.22} / $.45 \pm .08$ &  \textbf{.23} / $.15 \pm .01$ \\
		
		\textbf{ACL-17$\ddagger$} & $.78$ / $ .76 \pm .06$ & .06 / \textbf{.11} $ \pm .04$ \\ 
		
		\textbf{CONLL-16$\dagger$} & $.78$ / $ .77 \pm .04$ & .08 / $ .08 \pm .00$ \\
        
		\textbf{F1000-22$\ddagger$} & $.41$ / \textbf{.21} $ \pm .01$ &  $.19$ / \textbf{.41}$ \pm .04$ \\
		\bottomrule
	\end{tabular}
 }
	\caption{RSP performance: baseline (mean-score) / median $\pm$ standard deviation. MRSE lower is better, F1 higher is better. Reporting best performing LLM; $\dagger$ = RoBERTa, $\ddagger$ = BioBert; full results in Appendix \ref{as:experiments}.}
	\label{t:rsp}
\end{table}

\subsection{Pragmatic Labeling} The core function of peer review is to assess and suggest improvements for the work at hand: a good review summarizes the work, lists its strengths and weaknesses, makes requests and asks questions. Although some venues employ structured review forms to encourage review writing along these dimensions, issues of imbalanced feedback (e.g. focus only on weaknesses) \cite{ampere}, inconsistent uses of review sections, and lack of guidance for free-form reviews persist. Here, a pragmatic labeling model can provide feedback to reviewers to potentially revise, balance or rearrange their review report. Following up on prior studies on argumentative and pragmatic labeling schemata for free-form peer reviews \cite[etc.]{ampere, kuznetsov2022revise}, we use \corpus\ to explore the connection of pragmatic labels and structured review forms for the purpose of this assistance scenario.

\paragraph{Task} We cast pragmatic labeling as a sentence classification task, where given a review sentence $s^R_i$, the model should predict its pragmatic label $c_i \in C$. For this experiment we use the data from two distinct reviewing systems: F1000Research employs free-form full-text reviews; a subset of F1000-22 has been manually labeled with sentence-level pragmatic labels in the F1000RD corpus \cite{kuznetsov2022revise}. ACL Rolling Review, on the other hand, uses structured review forms split into sections, which we align to the F1000RD classes: \texttt{Strength}, \texttt{Weakness}, \texttt{Request} (\texttt{Todo} in F1000RD) and \texttt{Neutral} (\texttt{Recap}, \texttt{Other} and \texttt{Structure} in F1000RD). See Appendix \ref{as:experiments} for more details.

\paragraph{Setup} %
For F1000RD we consider all $5$k labeled sentences mapped to the respective classes. For ARR-22, we take review sentences longer than five characters (to filter out splitting errors), and label them by their respective review section, resulting in around $14$k labeled sentences. For each dataset, we split the instances at random disregarding their provenance to maximize the diversity across splits. 

\paragraph{Results} Table \ref{t:pragmatics} presents the pragmatic labeling results in- and cross-dataset. Expectedly, neural models outperform the majority baseline in-dataset by a large margin. Yet, cross-dataset application also yields non-trivial results substantially above the in-dataset majority baseline, despite the domain shift between the ARR-22 and F1000RD data. This has important implications, as it suggests that data from structured reviewing forms (entered by reviewers as part of the reviewing process) can be used to train free-text pragmatic labeling models, thereby significantly reducing the annotation costs. Nevertheless, the gap to the in-distribution supervised model remains, constituting a promising target for the follow-up work, that would need to disentangle the effects of domain, task and reviewing system shift on pragmatic labeling performance.

\begin{table}[t]
	\centering
	\begin{tabular}{r c c}
	    \toprule
	     & \textbf{ARR-22} & \textbf{F1000RD} \\
		\midrule
		\textbf{ARR-22$\dagger$} & $0.72 \pm 0.01$ & $0.50 \pm 0.01$ \\
		\textbf{F1000RD$\dagger$} & $0.54 \pm 0.01$ & $0.87 \pm 0.01$ \\
		\midrule
		\textbf{baseline (maj.)} & $0.11$ & $0.152$ \\
		\bottomrule
	\end{tabular} 
	\caption{Pragmatic labeling performance using $\dagger$RoBERTa. Rows: training. Columns: test. Macro-F1, median $\pm$ standard deviation, \textit{vs} majority class baseline.}
	\label{t:pragmatics}
\end{table}

\subsection{Guided Skimming for Peer Review}
Review writing typically requires multiple passes over the paper to assess its contents. Different paper types (e.g. dataset or method papers) require different reviewing strategies \cite{rogers-pr}; hence, the regions that require most scrutiny and rigor during reading vary across papers. Suggesting passages most relevant to the required reviewing style could encourage higher quality of reviewing and serve as a point of reference to junior reviewers.
We model this scenario via the novel \emph{guided skimming for peer review} task, in line with \citet{Fok2022ScimIF} who integrate passage recommendations into reading environments, reporting improved reading performance.

\paragraph{Task} We model the task as follows: given a paper, the model should rank its paragraphs $\textit{par}^P$ by relevance to the critical reading process. The training data for this task is derived from \emph{explicit links}, e.g. mentions of line numbers or sections in the reviews, which can be reliably extracted from reports in a rule-based fashion (see \ref{ass:skimm} for details) and used to draw cross-document links between review report sentences $s^R$ and the paragraphs of the papers they discuss $\textit{par}^P$ \cite{kuznetsov2022revise}\footnote{We opt against \textit{implicit links}, i.e. links without explicit markers, as their automatic detection remains challenging.}. Paper paragraphs with incoming explicit links are then considered "review-worthy", and the task is to rank such paragraphs above others for previously unseen papers with no available review reports. While the resulting task is a simplification of the actual skimming process during peer review, it is a first step towards exploring review-paper-links for modeling actual reviewer focus. We encourage future work to follow-up on this line of inquiry.

\paragraph{Setup} We use the ARR-22 and ACL-17 datasets, as they are of sufficient size and offer line numbers in the paper drafts. 
We extend and apply the explicit link extractor proposed by \citet{kuznetsov2022revise}, resulting in a total of $229$ papers with $743$ relevant passages with an average of $3.24 \pm 3.2$ of linked passages for ARR-22 and $87$ papers, $308$ passages and $3.54 \pm 3.13$ linked paragraphs per paper for ACL-17. We fine-tune LLMs using a binary classification objective, batching linked and unlinked paragraphs of the same paper together and rank by the output softmax of the positive score. We compare to a random baseline. Datasets are split by paper while ensuring a similar distribution of passages per paper across splits (see app. \ref{as:experiments}).

\begin{figure}[t]
	\centering
	\includegraphics[width=\linewidth]{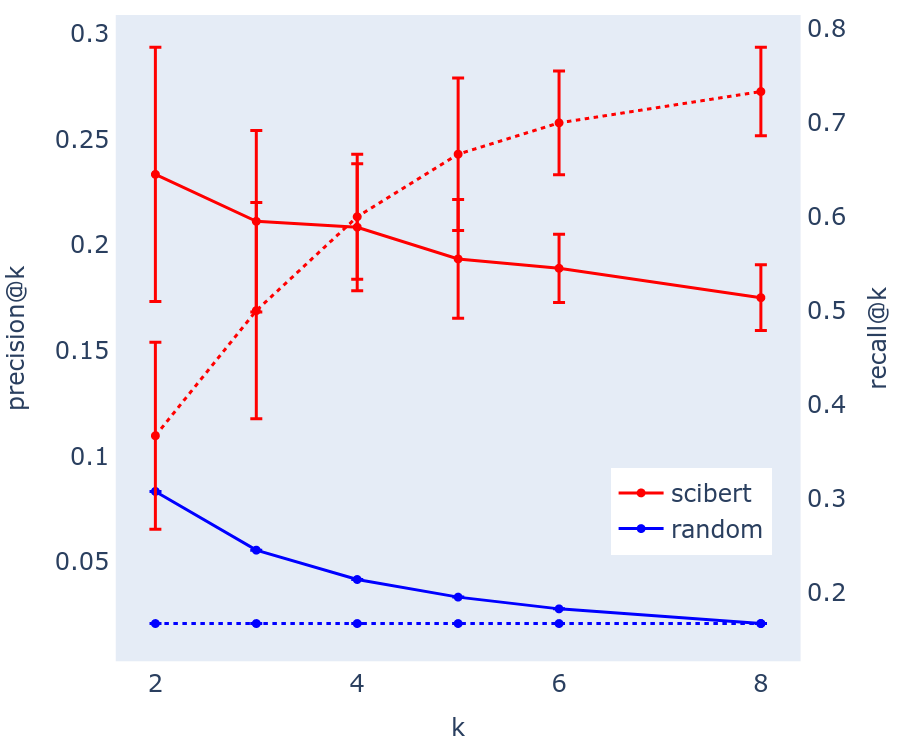}
	\caption{Precision (left) and recall (right, dotted lines) at k for SciBERT and random baseline on ACL-17.}
\label{fig:skimm}
\end{figure}

\paragraph{Results} \label{ss:skimm_res}
Figure \ref{fig:skimm} summarizes the Precision and Recall at $k$ for the best performing SciBERT model and the random baseline on ACL-17. Both recall and precision exceed the random baseline by a large margin and for all considered $k$. At around $k = 3$ roughly $50\%$ of the relevant passages are retrieved; while at the same rank around $21\%$ of the retrieved paragraphs are actually linked by the reviewers.
The mean reciprocal rank (MRR) measures the average position of the first relevant result within the rankings. SciBERT achieves an MRR of $0.41 \pm 0.05$ on ACL-17 and $0.34 \pm 0.03$ on ARR-22, outperforming the random baseline by $0.23$ and $0.18$, respectively. Overall, the LLM perform substantially above random despite discarding all context information of a paragraph (see appendix \ref{as:results} for a detailed analysis).
While \textit{guided skimming} is a non-trivial task, the above-random performance of the given LLM baseline shows promise for further research in this direction, which could also include an in-depth study considering the context of paragraphs -- e.g. their position in the logical structure of the paper.

\section{Further Applications}
\corpus\ includes rich representations and meta-data for a large number of inter-linked peer review and manuscript texts. This enables new NLP studies for peer reviewing assistance and beyond. In this section we highlight and critically reflect on further applications of \corpus\ in general NLP research and in practice.

Peer review reports are expert-written texts that typically reflect deep understanding of the underlying paper. Hereby, they can serve as a valuable resource for distant or direct supervision of general-purpose machine learning models. For instance, paper summaries in review reports can be used as a basis for a challenging document summarization task; aspect score disagreements between reviews may serve as a weak supervision signal for aligning arguments for and against the paper across reviews.

On the other hand, \corpus\ and peer reviewing data in general might be exploited to train models with dual uses and practical risks. 
One such example is automatic review generation, where given a paper the model should generate a review report. While a resulting model may be used benevolently as a paper-writing aid or an analytical device for the study peer review, using it to avoid the reviewing effort or replace the reviewer bears a wide range of risks. As it is implausible that current state-of-the-art NLP models could produce a non-generic, meaningful review of a novel paper \cite{reviewadvisor}, such applications could compromise the academic quality assurance process. Due to the central role of peer review in research and publishing, we emphasize that future work should carefully reflect on the real-world impact of any models developed based on \corpus.

\section{Conclusion and Intended Use}

We have presented \corpus -- the first clearly licensed large-scale reference corpus for the study of NLP for peer review. \corpus\ opens many new opportunities for the empirical study of scholarly communication. For NLP, it allows developing new annotated datasets based on clearly licensed, richly formatted, unified corpora that span multiple research domains, reviewing systems and time periods. It can be used as a testing ground for domain transfer \cite{gururangan, chronopoulou}, and enables the study of cross-document relationships between papers, paper revisions and peer reviews \cite{kuznetsov2022revise}. From the meta-scientific perspective, it allows comparing peer reviewing practices across research communities, and can provide crucial insights into how researchers review and revise scientific texts. Our task implementations and results can guide the development of NLP assistance systems and can be used for systematically comparing pre-trained language models in the context of peer reviewing applications. Finally, our resource can serve as a blueprint for future aggregate resources for the study of peer review in NLP.

\section*{Acknowledgements}
We would like to thank all parties that supported and advised us during the realization of the peer review data collection at ACL Rolling Review. ARR-22 would not have been possible without the discussion and approval by the ACL Committee on Reviewing in 2021 chaired by Hinrich Schütze, and we are grateful to everyone who reached out to us during the pilot stages of the project to make suggestions and express their concerns. We thank the editors-in-chief and the technical team of ACL Rolling Review for their support during the data collection; with a special thanks to Amanda Stent, Sebastian Riedel and Goran Glavaš. Additionally, we would like to thank the OpenReview.net team for their support during the implementation of the data collection at ARR. We express our gratitude to Nuria Bel for her feedback during the first iterations of our data collection initiative at COLING-2020, and to Richard Gerber for helping us with early technical challenges in SoftConf. Last but not least, we would like to thank our reviewers for their valuable suggestions, as well as the community members who have engaged in a lively debate on this initiative and provided us with both encouragement and useful feedback.

This research work has been funded by the German Federal Ministry of Education and Research and the Hessian Ministry of Higher Education, Research, Science and the Arts within their joint support of the National Research Center for Applied Cybersecurity ATHENE. It is co-funded by the European Union (ERC, InterText, 101054961). Views and opinions expressed are however those of the author(s) only and do not necessarily reflect those of the European Union or the European Research Council. Neither the European Union nor the granting authority can be held responsible for them. Finally, parts of this work have been co-funded by the German Research Foundation (DFG) as part of the PEER project (grant GU 798/28-1). This work is part of the InterText initiative\footnote{\url{https://intertext.ukp-lab.de/}}.

\section*{Limitations} \label{s:limits}

While we hope that our approach to data collection can serve as a benchmark for future NLP studies beyond peer review, we deem it equally important to explicitly outline the potential risks and limitations of \corpus{} and NLP for peer review in general. Our discussion below encourages future research in ethics and applied NLP for peer review; many of our considerations are not specific to peer review and are equally relevant to the applications of NLP in general.

From the data perspective, we deem it important to clearly state what \corpus\ is \textbf{not meant for}. Our data collection campaigns for ARR-22 and COLING-20 included an explicit disclaimer on the risks of author profiling on the peer reviewing data; we stress that such applications \textbf{violate the intended use of \corpus}. Furthermore, \corpus{} enables a wide range of new NLP assistance tasks for peer review. Yet, we encourage future studies of NLP for peer review to reflect carefully about the \textbf{potential risks and benefits} of new task definitions atop of peer reviewing data in general and \corpus{} specifically. For instance, the full automation of peer review, i.e. the generation of review reports given a paper, bears risks and dual uses.

Considering \textbf{diversity} in NLP datasets, we stress that even \corpus\ only covers a fraction of peer reviewing across all fields of science, and more data needs to be collected to enable fully representative NLP-based study of peer review. Due to the genre standards of scientific publishing our dataset only covers papers and reviews in \textbf{English language}. Multilingual scholarly document processing is overall poorly represented in NLP, and constitutes a promising avenue for future research. While our resource contains data from a wide range of domains, \textbf{research in arts and humanities is under-represented} due to the poor data availability. The trend towards open science and the adoption of responsible data collection practices \cite{workflow_3y} might bring reviewing data from previously unexplored domains and languages into NLP. We stress that any direct comparison based on our corpus would need to take into account reviewing practices and guidelines adopted by the respective communities.
Specifically, potential biases resulting from the donation-based collection for ARR-22 and COLING-20 should be taken into account.

From the task side, we highlight that implementations and resulting models presented here are meant to \textbf{exemplify} the proposed tasks, determine their technical feasibility, and \textbf{serve as a starting point for developing future NLP for peer review assistance systems}. As such, the provided implementations have \textbf{limitations}: for example, sentence-level pragmatic labels derived from structure-based ARR forms might contain noise since ARR forms group text on section level; guided skimming does not make use of implicit links, and explicit links are mostly based on line numbers and quotes, limiting the recall. Since we did not perform extensive hyperparameter search and tuning of the models, our results should \emph{not} be interpreted as a claim towards superiority of a particular model, approach or reviewing system.

We highlight that high intrinsic task performance does not necessarily translate into the extrinsic utility of NLP support in real-world reviewing environments. We thus deem it crucial to study the factors that affect the success of NLP assistance for peer reviewing. This includes the study of the human-machine interaction dynamics and its desiderata; for example, review score recommendations should be accompanied by explanations. We encourage extensive research on \textbf{risks of biases and errors} in NLP assistance models; for instance, a review score prediction model might learn undesirable biases against certain types of papers. Review writing assistance implemented in a real reviewing system should always be accompanied by carefully designed guidelines and policies.

Finally, we invite the community to reflect on the \textbf{potential societal consequences} of the individual NLP assistance tasks, \emph{even if NLP models accomplish them well}. To provide an example, our newly introduced guided skimming task assists during the effortful and time-intensive, yet crucial step of reading the paper under review. 
Although the guided skimming for peer review task models an intermediate step during reading and is intended to serve as an additional point of reference during the iterated skimming steps of peer review, such a technology might encourage reviewers to read \textit{only} the paragraphs suggested by the model. We argue that this risk of "lazy reading" is independent of the technology at hand; a reviewer that is institutionally incentivized to perform reviews as quick as possible, may read a paper superficially and settle with heuristics for their assessment \cite{rogers-pr} regardless of assistance. A greater risk, however, may be imposed by potential biases and errors of a guided skimming model, which could distract less experienced reviewers. While recent work on skimming assistance in scholarly articles \cite{Fok2022ScimIF} suggests a mature and reflected interaction of users with highlight recommendations and possible errors, this needs a specific investigation for the use case during peer review. On the other hand, a critical reading model may serve as a useful point of reference to guide reviewers to employ more scrutiny on the parts of the paper appropriate for this specific paper type, which ultimately may improve reviewing quality. We assess that the opportunities provided by the introduced review assistance tasks outweigh the potential risks in general, yet highlight that a targeted study is necessary to substantiate this assessment.

\bibliographystyle{acl_natbib}
\bibliography{short}

\newpage
\appendix
\section{Corpus Creation Details}
\label{as:corpus}

\subsection{Overview} \label{as:creation}

\paragraph{Revision Matching} Our data model assumes that each paper is associated with at least one revision -- yet some of the existing peer review datasets only provide the submitted drafts. To remedy this, we augment ACL-17, CONLL-16 and COLING-20 with camera-ready versions by matching the accepted paper draft titles and abstracts against the ACL anthology\footnote{\url{https://aclanthology.org/}}.

For each of the papers in the mentioned datasets, we extracted the title and abstract either from the provided meta-data or the PDF. We then considered all papers of the respective conference in the ACL anthology and retrieved the top five entries according to the sentence BLEU\footnote{Using NLTK 3.7, see \url{https://www.nltk.org/}} of the title and abstract with the paper at hand. Exact matches were included without manual verification, for the others the authors checked if the papers plausibly align. When in doubt, we opted to not include the matched camera-ready version. 

\paragraph{Paper Parsing} For all datasets except F1000-22, the raw paper inputs are PDFs, which are processed using GROBID \cite{GROBID} and translated into the ITG data model \cite{kuznetsov2022revise} that captures the structural information (i.e. sections, subsections, etc.), linking information (i.e. citations and references) and layout information (i.e. lines and pages) of a document. For the datasets with line numbers in paper drafts (ACL-17, CONLL-16, ARR-22), we first remove line numbers for parsing and then match the line and page information heuristically to the processed papers. Although manual spot checks suggest a high quality of GROBID and ITG parsing, errors are inevitable and we provide papers both in raw and parsed form. To process F1000 Research XML papers, we extend and modify the existing parser provided in the ITG library.

\paragraph{Review Parsing}
The reviews are converted into a standardized format that supports structured and unstructured reviews, with and without scores. This means that for ARR-22, COLING-20, ACL-17, and CONLL-16 we include the rich set of review scores, and for ARR-22 the diverse review sections (strengths, weaknesses, etc.) remain intact.

\paragraph{Metadata and Versions}
\corpus\ includes all available revisions of a paper. For F1000-22, each paper may have multiple revisions. Other datasets include at least the paper draft and if the paper was eventually published a camera-ready version of the paper. We associate rich metadata with each paper version, including the extracted title, abstract, authors (for accepted papers), and, if available, information on the paper type.

\paragraph{Personal data check} To ensure that the published reviews pose no risks related to personal information, each peer review text in \corpus\ -- including the data adopted from prior work -- was additionally validated by at least one NLP expert from our group. The initial analysis has identified 25 potentially problematic cases, of which 10 were deemed relevant upon a second check. Most of these cases contained either anonymous OpenReview.net identifiers within the review texts or included notes to the area chairs within the review main body. Despite low risk of cross-linking this information, we opted to discard these potentially privacy relevant sentences from the respective reviews.

\subsection{COLING-2020 Collection} \label{as:coling20}

The data for the COLING-20 dataset was collected during the 28th International Conference on Computational Linguistics. Independent from the actual reviewing process, we asked authors and reviewers to donate their anonymized drafts and reviews, respectively. 

\paragraph{Workflow} After reviewing was completed, we reached out to authors and reviewers, asking them to consent to the research use of their data and to grant a CC0 license to their texts, with the the silence period of two years after COLING-20 acceptance decisions (October 2020). Reviewers and authors were informed about the risks of author profiling based on their provided textual artifacts. In total roughly $1500$ anonymous reviews and $150$ drafts were donated in this way. To adhere to the principles proposed by \citet{workflow_3y}, we only include reviews and paper drafts for which both authors and reviewers agreed to donation; hereby avoiding any possibility of leaking confidential research ideas from the papers.

\subsection{ARR-2022 Collection} \label{as:arr}

ACL Rolling Review (ARR) is the unified and continuous reviewing system of the Association for Computational Linguistics (ACL). Papers are submitted in regular intervals, the \textit{cycles}, receive reviews and a meta-review. In case of a positive meta-review, the paper becomes eligible for submission to any of the ACL conferences including the Annual meeting of the ACL and Empirical Methods for Natural Language Processing, where program chairs make the final acceptance decision. Otherwise, the paper is revised, resubmitted and typically reviewed by the same set of reviewers in a later cycle.

\paragraph{Workflow}
We followed the workflow with the exact same license transfer agreements proposed by \citet{workflow_3y} to collect peer reviewing data of the cycles September 2021 trough January 2022 covering papers later accepted at the annual meeting of the ACL (ACL 2022) and the annual meeting of the north-american chapter of the ACL (NAACL 2022). Independent from the reviewing process, reviewers were presented the option to donate all peer reviews of each cycle in bulk anytime during the reviewing period. After acceptance decisions for the conferences were released, we reached out to the authors of accepted papers roughly one month before the actual conference took place. Authors and reviewers were informed about the risks of author profiling and review release. The collected dataset contains reviews for the final draft of a paper, but none of the previous revisions. Some reviews included in the dataset are therefore revisions of previous reviews or may contain references to those.

\subsection{F1000-22 Creation}
F1000Research is an open post-publication reviewing platform covering articles from various fields, from clinical medicine to scientific policy research to R package development, as well as different article types, including case studies, literature reviews, research articles and code documentation. Publications are published \emph{prior to acceptance}, and then can be approved, rejected or approved-with-reservations by one several invited reviewers. Publications can have multiple versions; each version is accompanied by open peer reviewing reports, author responses and amendment notes. All data on F1000Research is provided under an open license (CC-BY) in an easy-to-process JATS XML format.

\paragraph{Workflow}
F1000 Research provides an official API for collecting peer reviews and articles\footnote{\url{https://f1000research.com/developers}}. We retrieved the index of articles and reviews in July 2022 and subsequently downloaded all articles with reviews for all versions and in XML, as well as PDF format. We discarded $34$ articles with invalid file formatting and roughly $2000$ articles that lacked reviews for the first version, as these indicate stale submissions. We extract meta-data, reviews and author responses from the article JATS XML files.

\subsection{Extended Datasets Analysis}
\begin{figure} 
	\centering
	\includegraphics[width=0.95\linewidth]{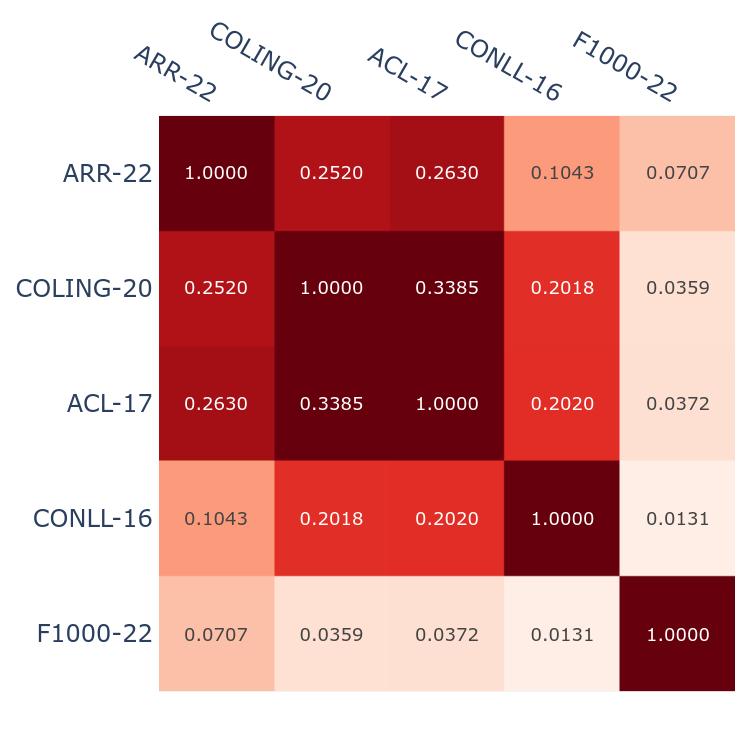}
	\caption{Vocabulary overlap on paper abstracts based on the Jaccard metric on lemmas of the datasets in \corpus.}
\label{afig:paper_vocab_overlap}
\end{figure}

We complement the domain overlap analysis between reviews by the same analysis on paper abstracts. The vocabulary overlap in Figure \ref{afig:paper_vocab_overlap} is generated under the same configuration (Jaccard metric on the top $10\%$ of the lemmas), but computed on paper abstracts for all datasets. We see that the vocabulary overlap in abstracts is very similar to the overlap in review texts. However, the absolute similarity values are overall lower. This supports the observation that reviews have a wider cross-domain shared vocabulary, while papers apparently employ a more specialized register.

\section{Experiment Details}
\label{as:experiments}

\subsection{Fine-tuning Setup} \label{as:finetuning}
The goal of our experiments is to identify the difficulty of each task and observe domain differences across datasets in \corpus. We therefore fine-tune well-established large language models (LLMs) on each of the tasks; while the training objectives and approaches vary, we omit extensive fine-tuning of the LLMs and instead focus on a base set of hyper-parameters close to the recommended ones. During a pilot study we observed relatively unstable training, which rendered extensive random hyper-parameter search infeasible for the scope of this work.

\paragraph{Fine-tuning}
Table \ref{at:hyperparams} depicts the different fine-tuning parameters per LLM used for all experiments unless stated otherwise. We use the huggingface transformers implementation of RoBERTa\footnote{\texttt{roberta-base}, \url{https://huggingface.co/roberta-base}} with roughly 125 million parameters, BioBERT\footnote{\texttt{dmis-lab/biobert-v1.1}, \url{https://huggingface.co/dmis-lab/biobert-v1.1}} with around 110 million parameters, and SciBERT\footnote{\texttt{allenai/scibert\_scivocab\_uncased}, \url{https://huggingface.co/allenai/scibert_scivocab_uncased}} with roughly 110 million parameters. We follow recommendations \cite{mosbach2020stability} for fine-tuning LLM on comparatively small datasets. For all of the models and datasets, we allow up to 20 epochs of training, employ a linear warm-up schedule (for $6\%$ of the training steps) with non-bias weight decay of $0.1$ and have $10$ repeated measures on different random seeds to account for different random initializations of the task-specific classifier heads. 

We implement the training and testing pipeline in pytorch lightning\footnote{1.7.7, \url{https://www.pytorchlightning.ai/}} using huggingface transformers\footnote{4.22.2, \url{https://huggingface.co/docs/transformers/index}}. For each run, we select the model with the best performance on the validation set at the end of each epoch. We implement an early stopping mechanism that stops fine-tuning if no improvement is observed after at most 8 epochs.

\begin{table}[t]
	\centering
	\begin{tabular}{r c c c}
	    \toprule
	     & \textbf{Learning Rate} & \textbf{Batch Size} &  \\
		\midrule
		\textbf{RoBERTa} & $2.00\times10^{-5}$ & $16$ & \\
		\textbf{BioBERT} & $1.00 \times10^{-5} $  & $16$ & \\
		\textbf{SciBERT} & $3.00\times10^{-5}$ & $32$ & \\
		\bottomrule
	\end{tabular} 
	\caption{Hyper-paramters for fine-tuning the large language models consistent across tasks.}
	\label{at:hyperparams}
\end{table}

\paragraph{Repeated Measures}
For each task and language model, we apply the fine-tuning procedure described above using the test and development sets. We repeat fine-tuning including model selection in total 10 times for each model and task. In each fine-tuning run we vary the random seed influencing  the order of batches and randomly initialized weights of the model. We report the used random seeds within the code provided along the submission.

\paragraph{Stratified Splitting}
For the tasks review score prediction and guided skimming for peer review, we split the datasets with a special stratification criterion. For review score prediction, we require that the distribution of reviews per paper is similar across splits. For guided skimming, we make sure that the splits have a similar distribution of relevant paragraphs per paper. To achieve this, we employ sklearn's stratified, binary split function\footnote{\texttt{train\_test\_split} of sklearn 0.0 \url{https://scikit-learn.org}} while mapping the considered numerical stratification criterion to a discrete space by assigning the real numbers to buckets. To realize a split into three datasets, we realize repeated binary splits.

\subsection{Pragmatic Labeling}
For pragmatics labeling we map the semi-structured review form of ARR and the labels of the F1000RD dataset \cite{kuznetsov2022revise} to the same set of labels. The mapping of labels is summarized in Table \ref{at:labels}. We highlight that, unlike the manually curated labels of F1000-RD, the sentences of ARR-22 are just extracted from the review forms which do not enforce full consistency with the respective section implying certain levels of noise in labels. For instance, some reviewers do mention strengths in the summary section of a review. Hence, our experiments are also targeted towards determining if this scalable approach to acquire a supervision signal is feasible; a further, detailed analysis of the quality of labels is a promising future direction of research. 

\begin{table*}[t]
	\centering
	\begin{tabular}{r c c c c}
	    \toprule
	     \textbf{Label} & \textbf{ARR Review Section} & \textbf{ARR-22} & \textbf{F10000-RD} & \textbf{F1000-22} \\
		\midrule
		\texttt{Neutral} & Summary & 3466 & Recap, Other, Structure & 1932 \\
        \texttt{Strength} & Strengths & 2565 & Strength & 415 \\
        \texttt{Weakness} & Weaknesses & 4458 & Weaknesses & 811 \\
        \texttt{Request} & Questions and Requests & 4257 & Todo & 1514 \\
        & & \textbf{14746} & & \textbf{4672} \\
		\bottomrule
	\end{tabular} 
	\caption{Pragmatic label mapping unifying F1000-RD and the ARR review form sections. The column ARR-22 and F1000-22 indicate the number of sentences with the given label within the respective datasets. The bottom row shows the total amount of sentences within each of the datasets.}
	\label{at:labels}
\end{table*}

\subsection{Guided Skimming for Peer Review} \label{ass:skimm}
We formulate guided skimming as a ranking task on the paragraphs of a paper considering their relevance to the writing of a peer review. To approach this task, we exploit explicit links from the review reports to the paper indicating that reviewers discuss a specific paragraph. 

\paragraph{Explicit Link Detection}
\citet{kuznetsov2022revise} propose a regular-expression-based algorithm to detect anchors (i.e. explicit mentions of structural elements in the paper) in review reports. The authors report an F1 score of $0.77$ (with a precision of $0.81$) for anchor identification and $0.64$ (with a precision of $0.66$) for their approach at matching explicit anchors to regions in the paper compared to human annotations. We conclude that the extraction of explicit links works reasonably well to be used as a proxy for reviewers' focus regions in the paper. Manual checks support this observation, in particular we see that explicit links seem to be identified with high precision, but comparatively low recall. Consequently, the resulting labels do contain certain levels of noise and reflect only a subset of the regions of the paper that are actually discussed in the reviews. We also highlight that implicit links in reviews, i.e. discussions of paper aspects that do not use explicit identifiers for paper regions, are not considered, as they are not readily available at scale.

Table \ref{at:regex} shows the regular expressions used for the extraction of explicit anchors in reviews. We extend the existing set of rules by line numbers and formulas as supported in ARR-22 and ACL-17. For matching these, we rely on the layout information extracted heuristically from the PDFs allowing a reliable mapping of paragraphs to line ranges. We aggregate the explicit links of all reviews for a paper to derive the set of focused paragraphs. We omit the frequency of links to paragraphs and consider only binary labels (linked or not-linked), to simplify the task and avoid making additional assumptions. While we in principle allow for any kind of explicit link (to paragraphs, sections, etc.) during extraction, we only include those that can be mapped to one specific paragraph leaving only quotes and line references in practice.

\begin{figure}
	\centering
	\includegraphics[width=\linewidth]{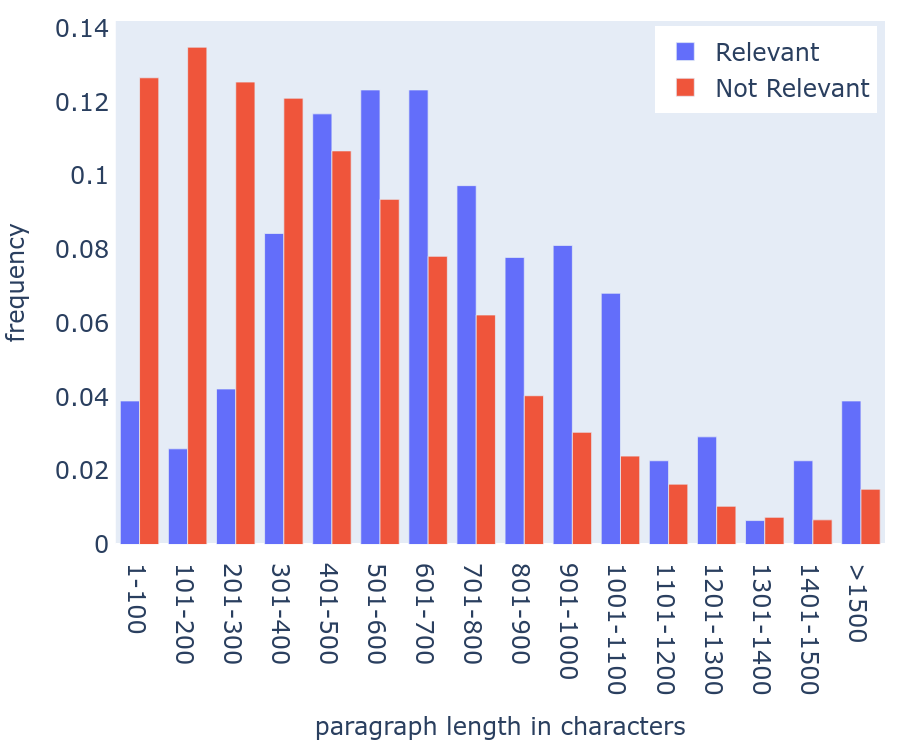}
	\caption{Distribution of paragraph lengths for ACL-17, where "relevant" marks paragraphs that were in fact referenced by reviewers.}
\label{fig:length_skimm_acl17}
\end{figure}

\begin{figure}
	\centering
	\includegraphics[width=\linewidth]{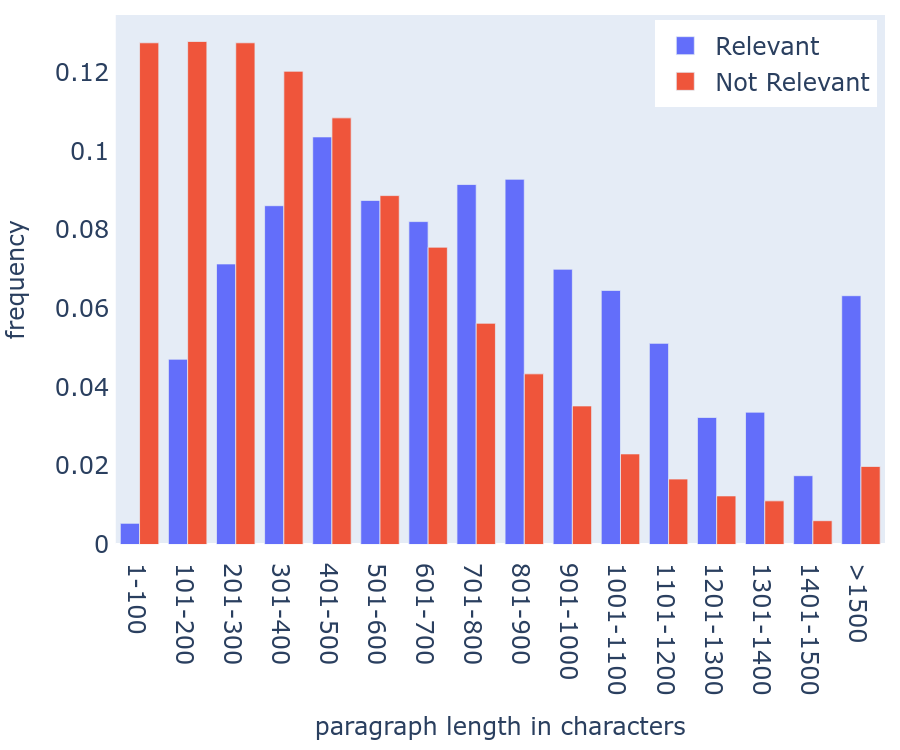}
	\caption{Distribution of paragraph lengths for ARR-22, where "relevant" marks paragraphs that were in fact referenced by reviewers.}
\label{fig:length_skimm_arr22}
\end{figure}

\paragraph{Relevant Paragraph Distribution within Papers}
In total the papers of ARR-22 have $12636$ paragraphs of which $1100$ (roughly $9\%$) are linked. In the ACL-17 dataset $600$ of $4642$ paragraphs ($13\%$) are referenced by reviewers explicitly.
We investigate the types of explicit links in each dataset. For ARR-22 $64\%$ of the relevant paragraphs are extracted from line-mentions in the reviews, while for ACL-17 these amount to $57\%$. Figures \ref{fig:length_skimm_acl17} and \ref{fig:length_skimm_arr22} show the histograms of text lengths for the relevant (i.e. linked) and not relevant (i.e. not linked) paragraphs of the papers. Overall, the length of relevant paragraphs tends to be higher than that of short paragraphs. We suspect that this phenomenon is encouraged by two interacting factors: first, paper parses are not perfect especially for splitting PDFs into structural elements leading to differently sized paragraphs in addition to naturally occurring size variations. Second, in combination with the first factor, longer paragraphs are simply more likely to be linked, because they have more content that can be discussed. This encourages a further line of work based on length-normalized paragraphs; we also investigate paragraph length as a spurious feature for the skimming task in the following.

Further on, we investigate the position of relevant paragraphs within the document hierarchy (sections, subsections, etc.) of the papers. For ARR-22 around half of the paragraphs originate from the level of sections rather than sub-sections, while for ACL-17 roughly two thirds lie on section level. This suggests no substantial bias towards a level in the document hierarchy.

\begin{figure}
	\centering
	\includegraphics[width=\linewidth]{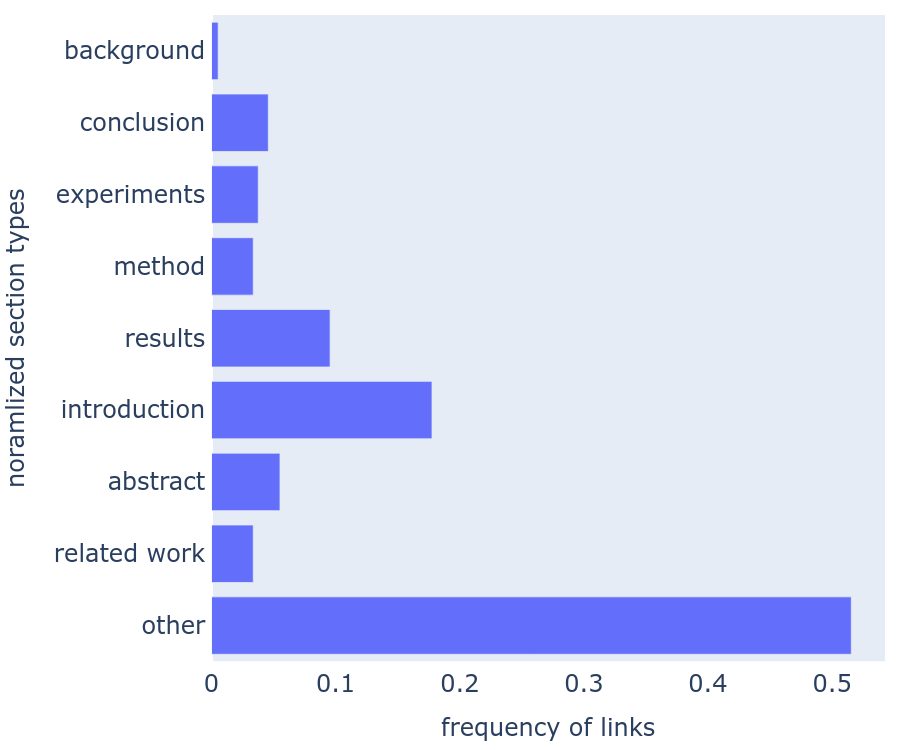}
	\caption{Distribution of links by normalized section type for ARR-22. Sections with non-standard names are mapped to the other category.}
\label{fig:sections_skimm_arr}
\end{figure}

\begin{figure}
	\centering
	\includegraphics[width=\linewidth]{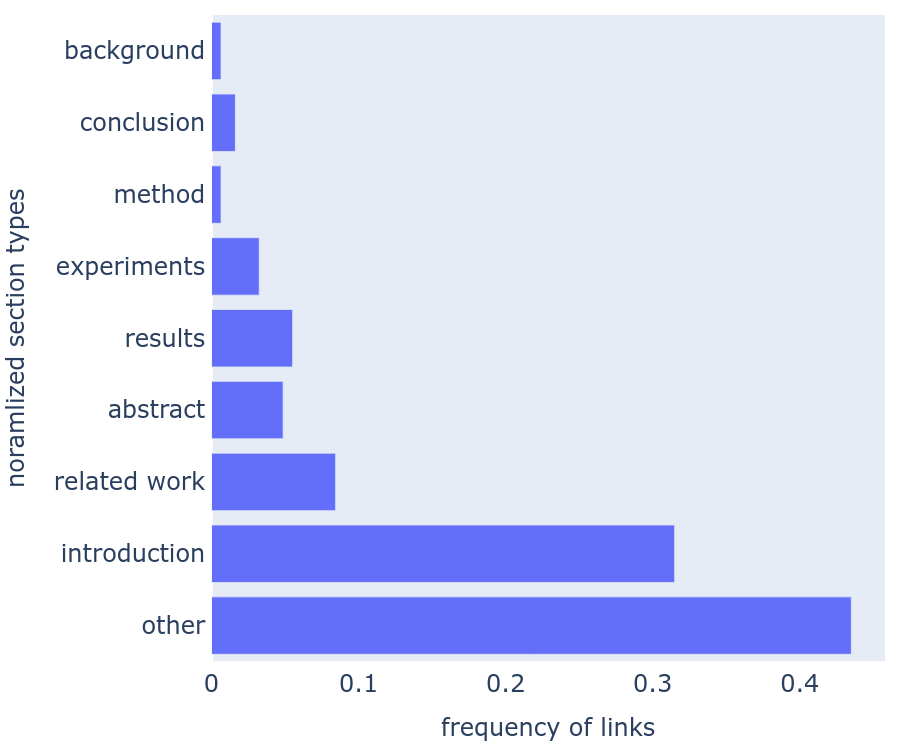}
	\caption{Distribution of links by normalized section type for ACL-17. Sections with non-standard names are mapped to the other category.}
\label{fig:sections_skimm_acl17}
\end{figure}

Finally, we analyse the distribution of linked paragraphs within a set of canonical sections typical for NLP papers, including e.g. introduction, method, and results. Figure \ref{fig:sections_skimm_arr} and \ref{fig:sections_skimm_acl17} show the frequency of links to each of these canonical section types. We map non-canonical section titles to the "other" category. In both datasets the number of explicit links pointing to the introduction is the highest when ignoring the section type "other". For ARR-22 this phenomenon is less pronounced, while the results section is slightly more often referenced than in ACL-17. Overall, there appear to exist natural areas of focus for reviewers as approximated by explicit links, but paragraphs of many different section types are in fact covered in both datasets. As our baseline models don't take structural information into account the observed light skews towards certain sections is unlikely to be the primary explanation for the non-trivial performance of the LLM reported in Section \ref{ss:skimm_res}.

\begin{table*}
	\centering
	\begin{tabular}{R{1.5cm} L{9cm} L{3.8cm}}
	    \toprule
	     \textbf{Type} & \textbf{Rule} & \textbf{Example}\\
		\midrule
        fig-ix & [Ff]igure | [Ff]ig(\textbackslash.?)  & \textit{Figure \underline{3} shows...}\\
table-ix & [Tt]able (\textbackslash.?) & \textit{I don't understand [...] mentioned in table \underline{1}.}\\
sec-ix &  [Ss]ection | [Ss]ec (\textbackslash.?) & \textit{There is a typo in the title of sec \underline{3}}. \\
sec-name & (?P<ix>[A-Z][A-z]+) | [“"'](?P<ix>[A-Z][A-z	]+)[”"'] [Ss]ection | (?P<ix>\textbackslash b[Tt]itle\textbackslash b)& \textit{In the \underline{"Methods"} section [...].}\\
quote & \textit{omitted due to excessive length -- including various forms of quotation marks} & \textit{The authors claim \underline{"..."}.}\\
ref-ix & [Rr]ef(\.|erence)?	?(?P<ix>\textbackslash d+) & \textit{ref \underline{23} is mal-formatted.} \\
page & (\textbackslash bp\textbackslash.?|page) & \textit{On \underline{page} 1 [...].} \\
paragraph & \textbackslash b(pp\textbackslash.?|paragraph|para\textbackslash.?)	?(?P<ix>\textbackslash d+) & \textit{paragraph \underline{3} introduces [...].}\\
        \midrule
line & \textbackslash b(line|l\textbackslash.?) ?(?P<ix>\textbackslash d+) | (?P<ix>first) | (?P<ix>second) | (?P<ix>third) | [IiOo]n	lines? (?P<ix>\textbackslash d+) | [IiOo]n l\textbackslash.? (?P<ix>\textbackslash d+) | [IiOo]n ll\textbackslash.? (?P<ix>\textbackslash d+) | \b(?P<ix>\d)(st|nd|rd) & \textit{On lines \underline{23-28} [...]}
\\
page & \textbackslash b(?P<ix>\textbackslash d)(st|nd|rd) | (?P<ix>first) | (?P<ix>second) | (?P<ix>third) & \textit{On page \underline{7} [...].}\\
formula & \textbackslash b[Ee]q(uations?|\.?)	?(?P<ix>\textbackslash d+) & \textit{There is an error in equation \underline{3}}\\
    	\bottomrule
	\end{tabular} 
	\caption{Selection of the regular expressions used for identifying explicit anchors in review reports. The horizontal line separates the rules proposed by \citet{kuznetsov2022revise} from the extended rules introduced in this work. The full set of rules is provided in the code associated with this work. For each type of explicit link, we provide an example.}
	\label{at:regex}
\end{table*}

\newpage
\section{Detailed Results} \label{as:results}

\subsection{Review Score Prediction}
We report the complete and more detailed results of the review score prediction models including all tested large language models (RoBERTa, BioBERT, SciBERT) and the mean score baseline in Table \ref{t:rsp_full}.

\paragraph{Additional Metrics}
All metrics are computed on the actual review score scale; hence model outputs are mapped back from the normalized scale. In addition to the mean root squared error (MRSE) and the F1-macro on discretized output predictions, we report the $R^2$ metric to measure the degree of fit of the regression models, as well as an additional score diversity criterion. For this criterion, we compute the overall score distribution for a model across all samples of the test set and compare it with the true, human score distribution (using KL-divergence) to measure whether the diversity of model scores aligns with human reviewers. A model that learns reviewers' rating behavior well, should also predict scores from a similar range of scores as humans do. Models over-focusing or ignoring single scores would have high scores; hence lower is better in this case.

\paragraph{KL-Divergence Interpretation}
We see that the mean overall score consistently shows the highest scores meaning the lowest similarity to the human score distribution. While the best model according to MRSE tends to have one of the lowest scores, this is not consistently true (e.g. RoBERTa on F1000-22). For COLING-20 and CONLL-16 the models perform on-par with mean baseline suggesting that they converged to predicting scores very close to the mean. Overall, even the best models still show notably different rating behavior from humans across all samples and datasets. 

\begin{table*}[t]
	\centering
	\begin{tabular}{r c c c c}
	    \toprule
	     & \textbf{MRSE $\downarrow$} &  \textbf{$R^2$ $\uparrow$} &\textbf{F1-macro $\uparrow$} & \textbf{Score KL-Divergence $\downarrow$}\\
		\midrule
		\textbf{ARR-22} &  & & \\
		RoBERTa & $0.37 \pm 0.02$&$0.16 \pm 0.05$&$0.46 \pm 0.04$&$0.28 \pm 0.11$ \\
		BioBERT & $0.45 \pm 0.03$&$-0.03 \pm 0.07$&$0.29 \pm 0.06$&$0.71 \pm 0.33$ \\
		SciBERT & $0.38 \pm 0.03$&$0.13 \pm 0.08$&$0.40 \pm 0.09$&$0.45 \pm -1.00$ \\
		mean score & $0.45$ & $-0.02$ & $0.24$ & $2.21$  \\
		\midrule
		
		\textbf{COLING-20} & & & \\
		RoBERTa & $0.45 \pm 0.08$ & $-0.33 \pm 0.24$&$0.15 \pm 0.01$&$1.74 \pm 0.07$ \\
		BioBERT & $0.44 \pm 0.09$&$-0.31 \pm 0.27$&$0.15 \pm 0.07$&$1.74 \pm 0.17$ \\
		SciBERT & $0.47 \pm 0.06$&$-0.38 \pm 0.16$&$0.15 \pm 0.01$&$1.74 \pm 0.09$ \\
		mean score & $0.22$ & $-0.06$ & $0.23$ & $1.74$  \\
		\midrule
		
		\textbf{ACL-17} &  & & \\
		RoBERTa & $0.79 \pm 0.04$&$-0.02 \pm 0.05$&$0.06 \pm 0.01$&$2.38 \pm 0.07$ \\
		BioBERT &	$0.76 \pm 0.06$&$0.02 \pm 0.07$&$0.11 \pm 0.04$&$1.76 \pm 0.34$ \\
		SciBERT & $0.78 \pm 0.07$&$-0.00 \pm 0.09$&$0.08 \pm 0.04$&$2.31 \pm 0.42$ \\
		mean score & $0.78$ & $0$ & $0.06$ & $2.38$  \\
		\midrule
		
		\textbf{CONLL-16} & & & \\
		RoBERTa & $0.77 \pm 0.04$&$0.01 \pm 0.05$&$0.08 \pm 0.00$&$2.81 \pm 0.00$ \\
		BioBERT & $0.87 \pm 0.04$&$-0.12 \pm 0.05$&$0.08 \pm 0.01$&$2.81 \pm 0.41$ \\
		SciBERT & $0.89 \pm 0.29$&$-0.15 \pm 0.38$&$0.08 \pm 0.12$&$2.81 \pm -1.00$ \\
		mean score & $0.78$ & $0$ & $0.08$ & $2.81$  \\
		\midrule
		
		\textbf{F1000-22} & & & \\
		RoBERTa &	$0.20 \pm 0.01$&$0.47 \pm 0.03$&$0.38 \pm 0.09$&$0.75 \pm 0.29$ \\
		BioBERT & 	$0.21 \pm 0.01$&$0.44 \pm 0.02$&$0.41 \pm 0.04$&$0.60 \pm 0.10$ \\
		SciBERT &	$0.21 \pm 0.01$&$0.45 \pm 0.02$&$0.40 \pm 0.06$&$0.65 \pm 0.18$ \\
		mean score & $0.41$ & $0$ & $0.19$ & $1.28$  \\
		\bottomrule
	\end{tabular} 
	\caption{MRSE is measured on non-normalized scores (i.e. absolute values cannot be compared across datasets); here, lower is better. F1-macro computed on discretized scores, where higher is better. KL-Divergence is computed relative to the human score distribution across all samples of each dataset; hence, a lower value indicates a more similar distribution compared to humans. We report the median score over ten runs and provide the standard deviation.}
	\label{t:rsp_full}
\end{table*}

\subsection{Pragmatic Labeling}
We extend the results presented in the main body of the paper by a concise error analysis for the within and out-of dataset experiments. In the following report the confusion matrices for the best performing LLM (RoBERTa) of the model that achieved the reported median performance.

\paragraph{Within-dataset Errors}
Table \ref{t:conf_f1000} shows the confusion matrix for RoBERTa on the F1000-RD dataset. We see the highest model confusion for the neutral class; strengths, weaknesses and requests are all most commonly confused with a neutral statement and vice-versa. We highlight that in our experiments the class neutral subsumes the more fine-grained neutral labels of F1000-RD like \texttt{summary} or \texttt{structure}, which might be one factor contributing to harded delineation.

In Table \ref{t:conf_arr}  we report the confusion matrix for RoBERTa on the ARR-22 dataset. Generally, we seem a similar pattern as for F1000-RD: the neutral class is most often confused with the others. As the review forms are not strictly enforced, it is likely that strengths and weaknesses are already reported in the summary section correlating to the \texttt{neutral} label encouraging this confusion. Likewise several neutral, factual sentences exist in the strength and weaknesses sections. This shows one limitation of using structure review forms as a proxy for review sentence pragmatic. Interestingly, we see that requests and weaknesses are very commonly confused. While the noisy supervision labels might again be a contributing factor, this aligns with the reported highest human disagreement for these two classes by \citet{kuznetsov2022revise}.

\begin{table*}[t]
	\centering
	\begin{tabular}{r c c c c}
	    \toprule
	     & \texttt{strength} & \texttt{weakness} & \texttt{request} & \texttt{neutral} \\
		\midrule
        \texttt{strength} &63&1&0&7\\
        \texttt{weakness} &3&144&5&14 \\
        \texttt{request} &4&6&270&6 \\
        \texttt{neutral} &11&31&23&346\\
		\bottomrule
	\end{tabular} 
	\caption{Confusion matrix of RoBERTa on F1000-RD. The entry in row $i$ and column $j$ denotes that the label $j$ was predicted by the model for a sample of true class $i$.}
	\label{t:conf_f1000}
\end{table*}

\begin{table*}[t]
	\centering
	\begin{tabular}{r c c c c}
	    \toprule
	     & \texttt{strength} & \texttt{weakness} & \texttt{request} & \texttt{neutral} \\
		\midrule
        \texttt{strength} &356&30&13&103\\
        \texttt{weakness} & 29&579&207&59 \\
        \texttt{request} & 30&254&565&22 \\
        \texttt{neutral} & 49&60&11&581\\
		\bottomrule
	\end{tabular} 
	\caption{Confusion matrix of RoBERTa on ARR-22. The entry in row $i$ and column $j$ denotes that the label $j$ was predicted by the model for a sample of true class $i$.}
	\label{t:conf_arr}
\end{table*}

\paragraph{Out-of-dataset Errors}
We inspect the errors of the best performing models trained on ARR-22 and tested on F1000-RD, and vice versa. Table \ref{t:conf_arr_f1000} reports the confusion matrix for the first case. We see that unlike the in-domain trained model on F1000-RD weaknesses are frequently confused with requests and vice versa. This confirms that the request and weaknesses sections in the ARR review forms lead to the most ambiguity in the supervision labels as already hypothesized in the previous paragraph. Similar to the in-domain trained model the neutral class is the hardest to predict for the model. Overall, the transfer performance lies in a promising range that suggests more efforts on few-shot and cross-domain transfer of models are good future directions of research.

The transfer of a model trained on F1000-RD to ARR-22 aligns with previous observations for the within evaluation on ARR-22: most confusion is observed for the neutral class. Additionally, the model predicts the class \texttt{request} very frequently for sentences belonging to \texttt{weaknesses}. This supports the hypothesis that many sentences in the weakness section of ARR are actually requests, as the model trained on F1000-RD is based off of human gold annotations.

\begin{table*}[t]
	\centering
	\begin{tabular}{r c c c c}
	    \toprule
	     & \texttt{strength} & \texttt{weakness} & \texttt{request} & \texttt{neutral} \\
		\midrule
        \texttt{strength} & 54 & 7 & 2 & 8\\
        \texttt{weakness} &  4 & 106 & 55 & 1 \\
        \texttt{request} & 2 & 75 & 209 & 0 \\
        \texttt{neutral} &  33 & 171 & 127 & 80\\
		\bottomrule
	\end{tabular} 
	\caption{Confusion matrix of RoBERTa trained on ARR-22 and transferred to F1000-RD. The entry in row $i$ and column $j$ denotes that the label $j$ was predicted by the model for a sample of true class $i$.}
	\label{t:conf_arr_f1000}
\end{table*}

\begin{table*}[t]
	\centering
	\begin{tabular}{r c c c c}
	    \toprule
	     & \texttt{strength} & \texttt{weakness} & \texttt{request} & \texttt{neutral} \\
		\midrule
        \texttt{strength} & 299 & 9 & 19 & 175\\
        \texttt{weakness} &  38 & 295& 290& 251\\
        \texttt{request} & 30& 160& 451 &230 \\
        \texttt{neutral} &  65 & 32 & 26 & 578\\
		\bottomrule
	\end{tabular} 
	\caption{Confusion matrix of RoBERTa trained on F1000-RD and transferred to ARR-22. The entry in row $i$ and column $j$ denotes that the label $j$ was predicted by the model for a sample of true class $i$.}
	\label{t:conf_f1000_arr}
\end{table*}

\subsection{Guided Skimming for Peer Review}

In addition to the at-k-measures and aggregate metrics of the skimming performance, as reported in the main body of this work, we provide the full results of all models and additional metrics in this section.

\paragraph{Paragraph Length Baseline}
As reported in \ref{ass:skimm}, the length distribution of linked and non-linked paragraphs might be a useful spurious feature for ranking the paragraphs by relevance to the guided skimming process. While the truncation of model inputs of the used large language models to $512$ tokens makes is unlikely that the models are at risk of exploiting sequence length to achieve non-trivial performance, future approaches using, for instance, the full paper text or full paragraphs might do so. Hence, in the following we also report the performance of the baseline that ranks paragraphs by their number of characters.

\paragraph{Ranking Measures}
Table \ref{t:skimm_full}  reports the performance of the LLMs and the baselines on ACL-17. We consider the mean reciprocal rank (MRR) and the area under the receiver operating characteristic curve (AUROC) as overall measures of the ranking quality in addition to the at-k-measures. For ACL-17, SciBERT shows the best performance according to MRR, AUROC and Precision@3. While the model performs substantially above the random baseline, the margin towards the length baseline is small, especially for the AUROC. Overall, the ranking produced by SciBERT leads to a higher precision in the top ranks, but seemingly performs on-par with the length baseline for the overall ranking performance. This shows the difficulty of the task at hand, but at the same time suggests that there is a useful training signal in the paragraph texts beyond their mere length.

The results on ARR-22 as shown in \ref{t:skimm_full_arr} are very similar. Although all models perform above random, the paragraph length baseline is hard to beat. Here, we can observe some improvements in terms of recall and precision in the top ranks, but the overall ranking performance lies below the length baseline.

\begin{figure}[t]
	\centering
	\includegraphics[width=\linewidth]{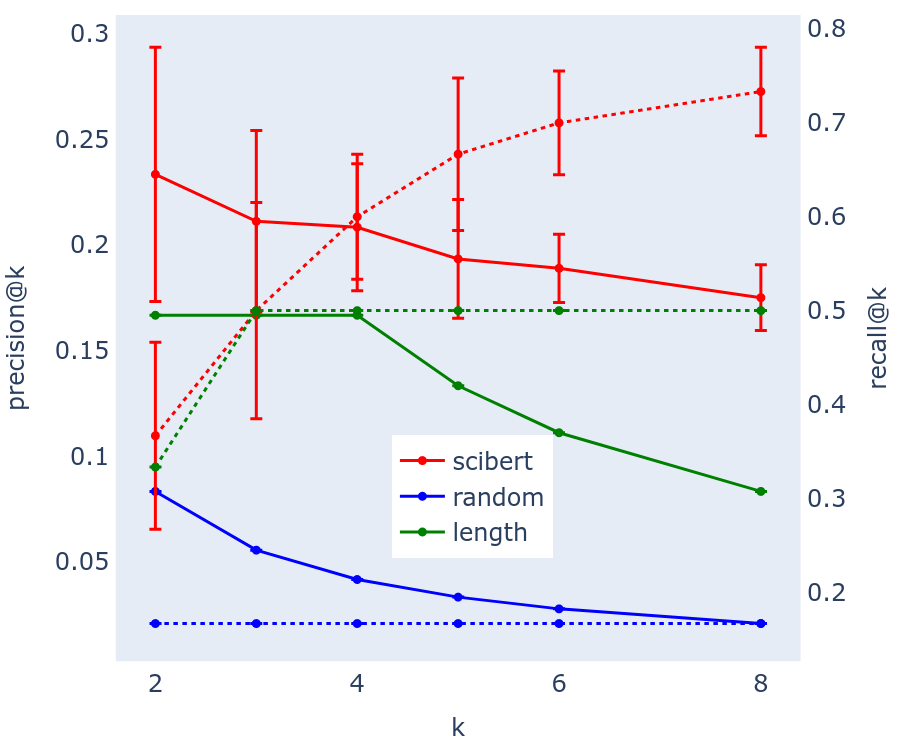}
	\caption{Precision (left) and recall (right, dotted lines) at different ranks k for SciBERT, random and length baseline on ACL-17.}
\label{fig:skimm_acl17_full}
\end{figure}

\paragraph{Conclusion}
The length of the paragraphs seems a relevant feature that might be exploited as a spurious decision criterion by the models. However, especially in the top ranks for ACL-17 (see \ref{fig:skimm_acl17_full}) the LLM seem to pick up information beyond mere paragraph length. We suspect that structural and contextual information would be beneficial for this hard task that would increase the margin towards the paragraph length baseline. Additionally, more elaborate training regimes considering list-wise losses a interesting future directions. To eliminate the risk of learning paragraph length as a spurious pattern the normalization of paragraph lengths by different segmentation techniques is promising.

\begin{table*}[t]
	\centering
	\begin{tabular}{r c c c c}
	    \toprule
	    & \textbf{MRR} $\uparrow$ & \textbf{AUROC} $\uparrow$ & \textbf{Precision @ 3} $\uparrow$ & \textbf{Recall @ 3} $\uparrow$\\  
		\midrule
		  RoBERTa& $0.46 \pm 0.05$& $0.73 \pm 0.06$& $0.22 \pm 0.03$& $0.53 \pm 0.05$ \\
BioBERT& $0.40 \pm 0.11$& $0.76 \pm 0.06$& $0.19 \pm 0.05$& $0.47 \pm 0.08$ \\
SciBERT& $0.43 \pm 0.06$& $0.77 \pm 0.04$& $0.21 \pm 0.04$& $0.50 \pm 0.12$ \\
Length Baseline& $0.34 \pm 0.00$& $0.76 \pm 0.00$& $0.17 \pm 0.00$& $0.50 \pm 0.00$ \\
Random Baseline& $0.14 \pm 0.05$& $0.47 \pm 0.06$& $0.08 \pm 0.03$& $0.20 \pm 0.08$ \\
		\bottomrule
	\end{tabular} 
	\caption{Performance on ACL-17. Mean Reciprocal Rank (MRR), area under the receiver operating characteristic curve (AUROC) and precision and recall at $k=3$ reported for the LLM, length baseline and random baseline. For all metrics, higher is better.}
	\label{t:skimm_full}
\end{table*}

\begin{table*}[t]
	\centering
	\begin{tabular}{r c c c c}
	    \toprule
	    & \textbf{MRR} $\uparrow$ & \textbf{AUROC} $\uparrow$ & \textbf{Precision @ 3} $\uparrow$ & \textbf{Recall @ 3} $\uparrow$\\  
		\midrule
RoBERTa & $0.30 \pm 0.04$ & $0.68 \pm 0.03$ & $0.15 \pm 0.03$ & $0.33 \pm 0.05$ \\
BioBERT & $0.35 \pm 0.02$ & $0.69 \pm 0.03$ & $0.17 \pm 0.03$ & $0.38 \pm 0.04$\\
SciBERT & $0.35 \pm 0.03$ & $0.70 \pm 0.02$ & $0.17 \pm 0.03$ & $0.40 \pm 0.05$\\
Length Baseline & $0.31 \pm 0.00$ & $0.75 \pm 0.00$ & $0.15 \pm 0.00$ & $0.28 \pm 0.00$\\
Random Baseline & $0.18 \pm 0.04$ & $0.49 \pm 0.02$ & $0.07 \pm 0.02$ & $0.20 \pm 0.05$\\
		\bottomrule
	\end{tabular} 
	\caption{Performance on ARR-22. Mean Reciprocal Rank (MRR), area under the receiver operating characteristic curve (AUROC) and precision and recall at $k=3$ reported for the LLM, length baseline and random baseline. For all metrics, higher is better.}
	\label{t:skimm_full_arr}
\end{table*}

\begin{figure}[t]
	\centering
	\includegraphics[width=\linewidth]{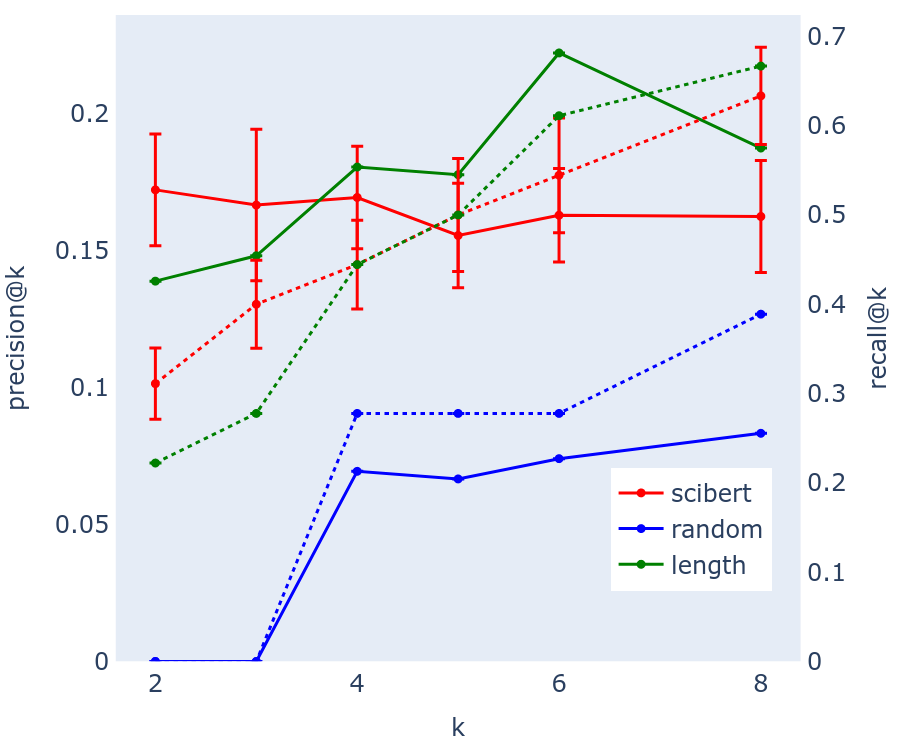}
	\caption{Precision (left) and recall (right, dotted lines) at different ranks k for SciBERT, random   and length baseline on ARR-22.}
\label{fig:skimm_arr_full}
\end{figure}

\end{document}